\documentclass{article}

\usepackage{spconf,amsmath}
\usepackage{graphicx}\graphicspath{ {images/} }
\usepackage{algorithm, algorithmic}
\usepackage{multirow, multicol, booktabs, microtype, xcolor}
\usepackage{balance}
\usepackage{hyperref}

\usepackage{enumitem}
\setlist{nosep, leftmargin=14pt}
\usepackage{booktabs}
\usepackage{tabularx}

\title{Prompting Medical Vision-Language Models to Mitigate Diagnosis Bias by Generating Realistic Dermoscopic Images}

\name{Nusrat Munia, Abdullah-Al-Zubaer Imran}
\address{Department of Computer Science\\
University of Kentucky, Lexington, KY 40506, USA}
\begin{document}

\maketitle
\begin{abstract}

Artificial Intelligence (AI) in skin disease diagnosis has improved significantly, but a major concern is that these models frequently show biased performance across subgroups, especially regarding sensitive attributes such as skin color. To address these issues, we propose a novel generative AI-based framework, namely, Dermatology Diffusion Transformer (DermDiT), which leverages text prompts generated via Vision Language Models and multimodal text-image learning to generate new dermoscopic images. We utilize large vision language models to generate accurate and proper prompts for each dermoscopic image which helps to generate synthetic images to improve the representation of underrepresented groups (patient, disease, etc.) in highly imbalanced datasets for clinical diagnoses. Our extensive experimentation showcases the large vision language models providing much more insightful representations, that enable DermDiT to generate high-quality images. Our code is available at {\color{purple}\href{https://github.com/Munia03/DermDiT}{https://github.com/Munia03/DermDiT}}.

\end{abstract}

\section{Introduction}

\label{sec:intro}
The rapid advancement of machine learning (ML) models in medical imaging has greatly improved the ability of diagnostic tools to detect malignancies in skin diseases. Recent research indicates that skin disease diagnosis models frequently show bias against certain subgroups \cite{tschandl2018ham10000,kinyanjui2020fairness,groh2021fitzpatrick}. These biases arise from sensitive attributes such as skin tone, age, and gender due to the underrepresentation of the subgroups in the training datasets. For example, diagnostic accuracy is usually poorer on images of darker skin tones compared to those with lighter skin tones \cite{ddi}. 

\begin{figure}[t]
    \centering
    \includegraphics[width=\linewidth, trim={3cm 0cm 0cm 0cm}, clip]{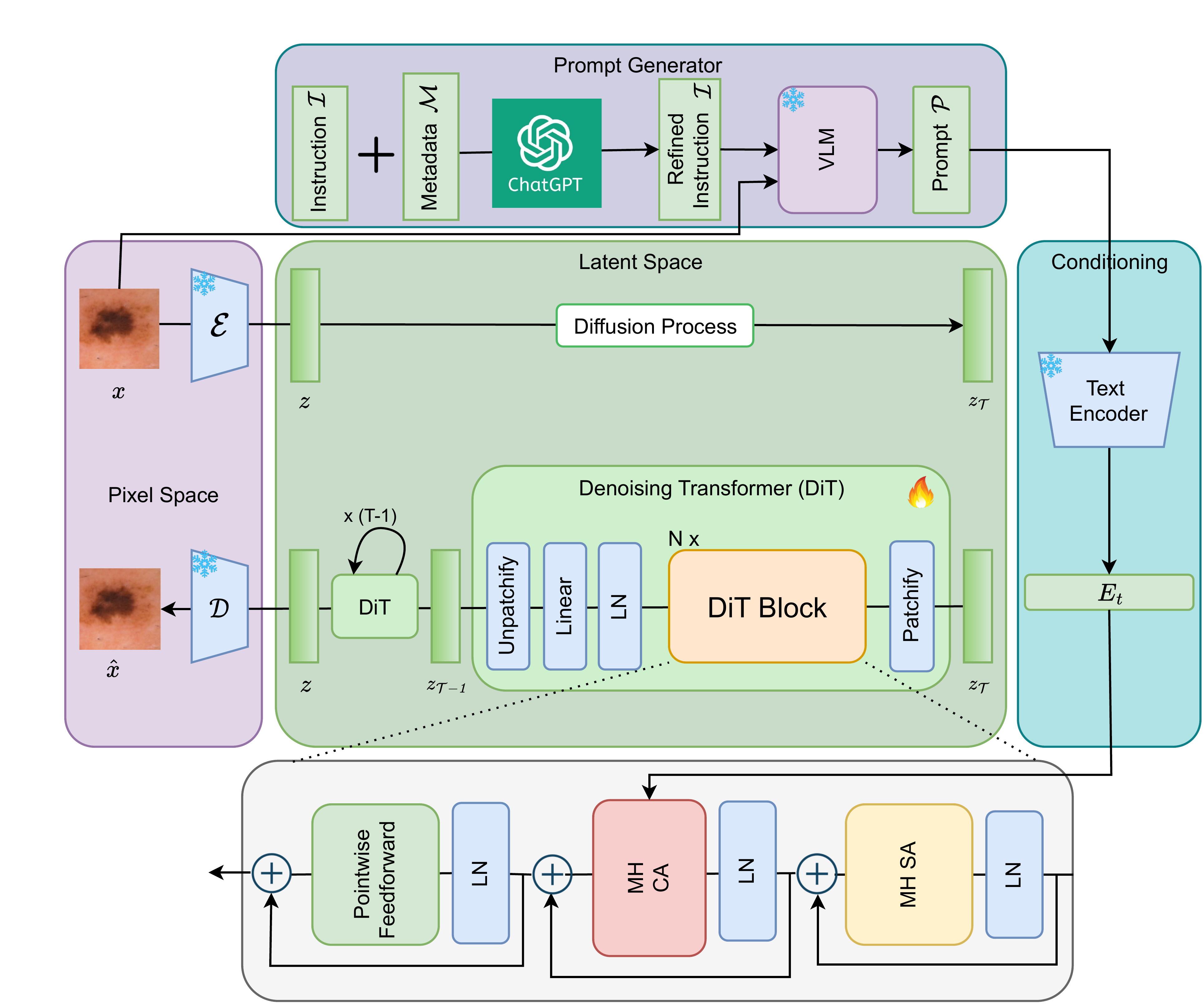}
    \caption{Proposed DermDiT: A diffusion transformer model generates new images conditioned on text prompts. These prompts are created by a Prompt Generator, which leverages a Vision-Language Model (VLM) to produce descriptive text based on input dermoscopic images and their associated metadata.}
    \label{fig:arch1}

\end{figure}

Several works are available in the literature to ensure fairness in artificial intelligence (AI)-based skin disease diagnosis \cite{xtranprune,gen_iterative,detecting_melanoma,fairprune,ddi,me_fairprune,fairadabn}. For example, two separate models are trained for lighter and darker skin tones and combined via an ensemble mechanism \cite{safe}. A de-biasing technique is proposed to unlearn skin tone features from the images to mitigate skin tone bias \cite{detecting_melanoma}. The FairAdaBN \cite{fairadabn} proposes adaptive batch normalization for sensitive attributes with a loss function to minimize the difference in prediction probability between subgroups. Pruning sensitive nodes from the model to make the model independent of sensitive attributes has been applied to mitigate unfairness \cite{xtranprune,fairprune,me_fairprune}. Although all these models improve fairness performance to some extent, there is still a performance gap due to the lack of a large diverse dataset. A diverse dataset, Diverse Dermatology Images (DDI) \cite{ddi}, is proposed for this, which is not large enough to train a deep learning model but can be used for evaluating the fairness of the diagnostic models.

In recent years, generative models like Generative Adversarial Networks (GANs) and Diffusion Models have gained popularity for generating images that resemble real datasets.  A GAN-based augmentation method is adopted for debiasing common artifacts like hair, ruler, frame, etc. without considering demographic attributes \cite{de_biasing2022}. However, Diffusion Models such as the U-Net-based Stable Diffusion model \cite{stable_diff} and transformer-based Diffusion Transformer \cite{dit} have shown significant performance improvements, particularly in generating high-quality images compared to GANs. The diffusion generative model is used to synthesize samples for under-represented groups during the training of the disease classification model \cite{gen_iterative}.
A conditional diffusion generative model, DermDiff, is proposed to generate new samples for under-represented groups to improve performance \cite{munia2025dermdiff}.  Diffusion models can be conditioned on class labels, text prompts, or others.  This condition can have a major impact on the performance of the generative models. In this work, we propose to leverage large vision language models to improve the performance of generative models.

Vision Language models (VLM) are achieving significant improvements in tasks like image captioning, visual question answering (VQA), and multimodal reasoning. CLIP (Contrastive Language–Image Pretraining) \cite{clip_text}, BLIP (Bootstrapping Language-Image Pretraining) \cite{blip}, and LLaVA (Large Language and Vision Assistant) \cite{llava} are gaining popularity for their multimodal data understanding and reasoning. As these models are trained on large vision and text data pairs, they generalize across a variety of tasks with minimal fine-tuning. There are some VLMs only specialized for the healthcare domain, MedCLIP \cite{medclip} similar to CLIP, and LLaVA-Med \cite{llava-med} similar to LLaVA, but these are trained on medical data only. These models focus on integrating medical images with text-based information to enhance clinical tasks such as medical diagnosis.
We propose to integrate VLMs with the DiT model to improve the performance of the generative model. Generating better synthetic images will lead to creating a balanced dataset for fair performance on disease diagnosis. Our specific contributions can be summarized as follows:
\begin{itemize}
    \item A generative model framework, DermDiT, conditioned on text prompts by integrating VLM
    \item A skin disease diagnosis model trained on synthetic images to perform fairly across different subgroups
    \item A balanced synthetic dataset including different attributes, especially focusing on minor subgroups.

\end{itemize}

\section{Methods}
\label{sec:methods}

Fig.~\ref{fig:arch1} illustrates the proposed DermDiT model. DermDiT is conditioned on text prompts generated via a VLM. It has two parts: the conditioning and the diffusion transformer.  

\subsection{Conditioning}
\textbf{Prompt Generator:} For a skin disease image $x$, we combine its metadata set $M = \{m_1, m_2, m_3,\dots,m_k\}$ where $m_1, m_2,..$ are attributes such as skin tone, gender, etc. We generate an instruction $I$ including all the attributes of the metadata, we further refine this instruction leveraging ChatGPT \cite{chatgpt}. We provide this instruction to the VLM (LLaVA or domain-specific LLaVA-Med) with the corresponding image to generate a descriptive caption of the input dermoscopic image $x$. The VLM model generates the image-specific text prompt ($p$)  given the instruction $I$, including all the attributes in the metadata information. 
\begin{equation}
p = VLM(I, x).
\end{equation} 

\noindent\textbf{Text Encoder:} A text encoder converts a text into its vectorized representation. We use a pre-trained text encoder such as CLIP \cite{clip_text} to get the text embeddings $E_t$ for the text prompt $p$. These embeddings represent the semantic information of $p$ in a high-dimensional space. CLIP is chosen as it learns a shared embedding space for both text and images, making it effective for aligning textual descriptions with visual content.
\begin{equation}
    E_t = CLIP (p).
\end{equation}
\subsection{Diffusion Transformer}
The diffusion model consists of two processes:  a forward process (diffusion), where noise is gradually added to data, and a reverse process (denoising), where noise is removed step-by-step to generate new data samples. 

\noindent\textbf{Forward process (Diffusion):}  For a dermoscopic image $x_0$ the forward process gradually adds noise to the image $x_0$: $q(x_t|x_0) = \mathcal{N}(x_t; \sqrt{\bar{\alpha}_t}x_0, (1 - \bar{\alpha}_t)\mathbf{I})$, where $\bar{\alpha}_t$ are hyperparameters. With reparameterization, we sample $x_t = \sqrt{\bar{\alpha}_t}x_0 + \sqrt{1 - \bar{\alpha}_t}\epsilon_t$, where $\epsilon_t \sim \mathcal{N}(0, \mathbf{I})$.

\begin{figure*}[t]
    \centering
    \resizebox{\textwidth}{!}{
    \begin{tabular}{cccccc} 

        \includegraphics[width=0.4\linewidth]{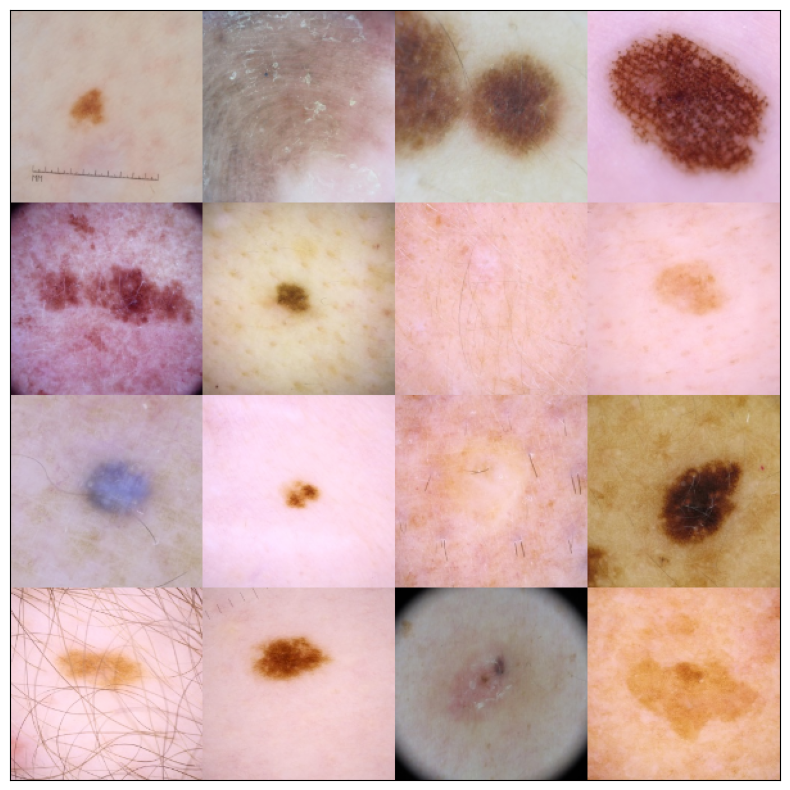} &
        \includegraphics[width=0.4\linewidth]{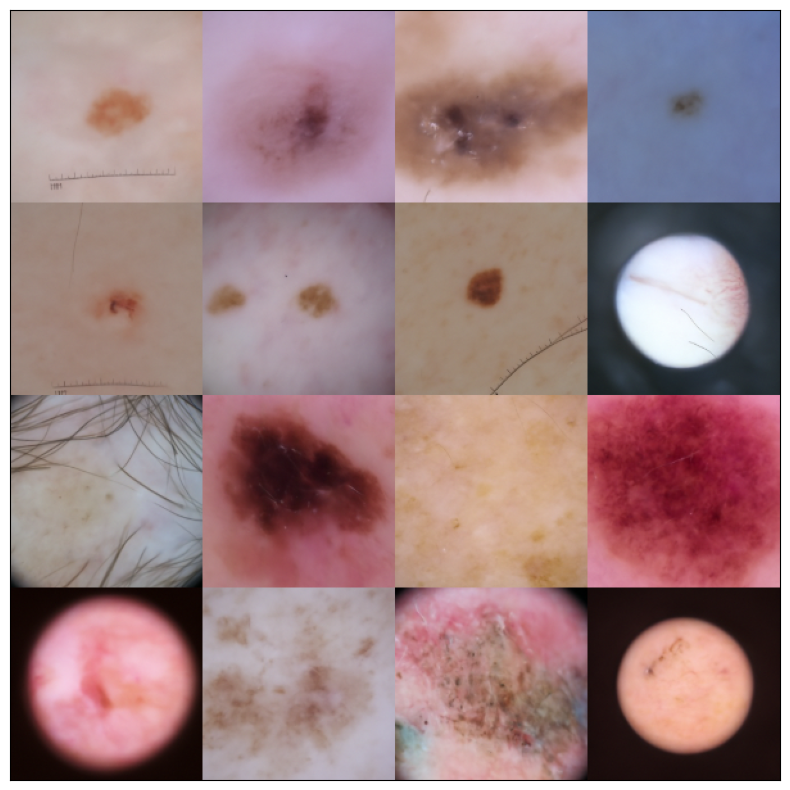} &
        \includegraphics[width=0.4\linewidth]{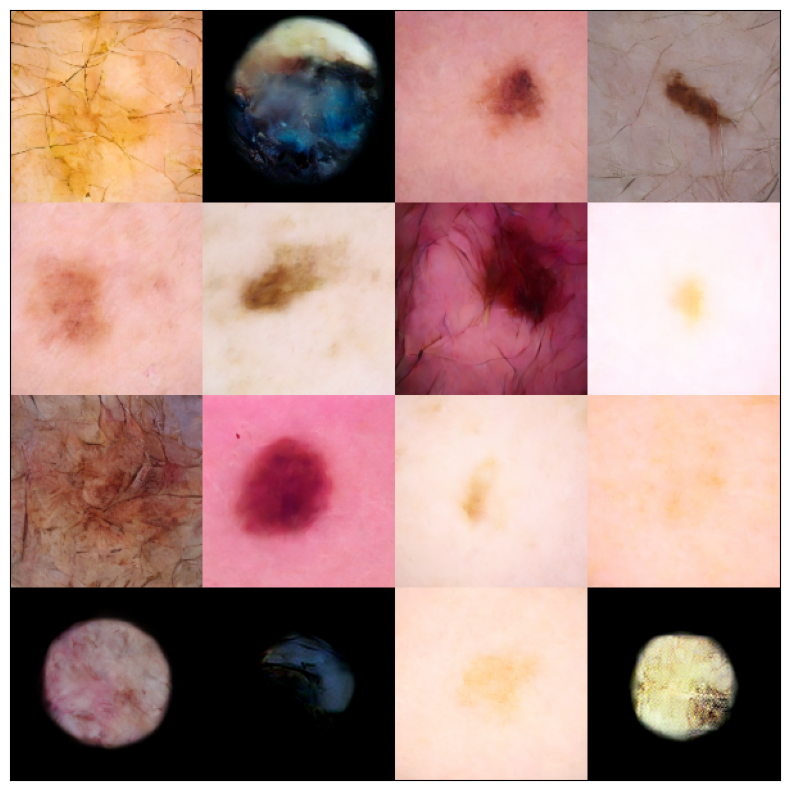} & \includegraphics[width=0.4\linewidth]{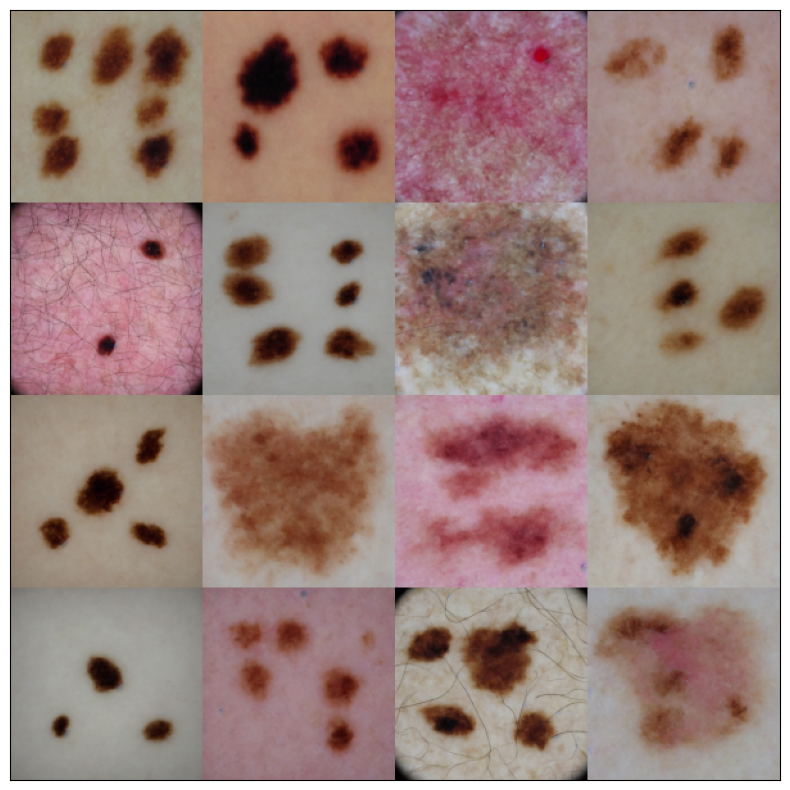}  & 
        \includegraphics[width=0.4\linewidth]{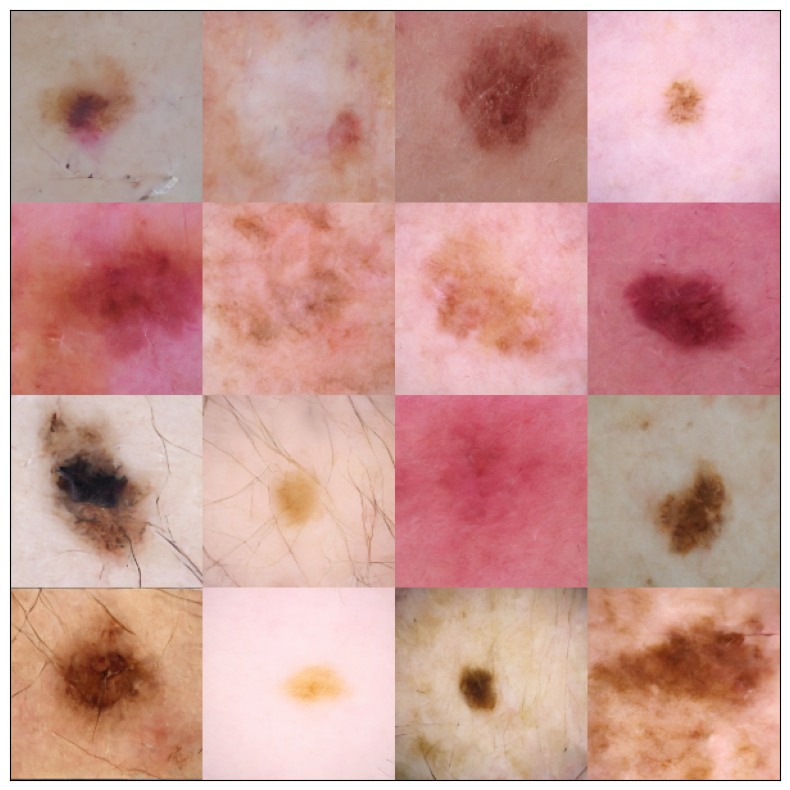} &
        \includegraphics[width=0.4\linewidth]{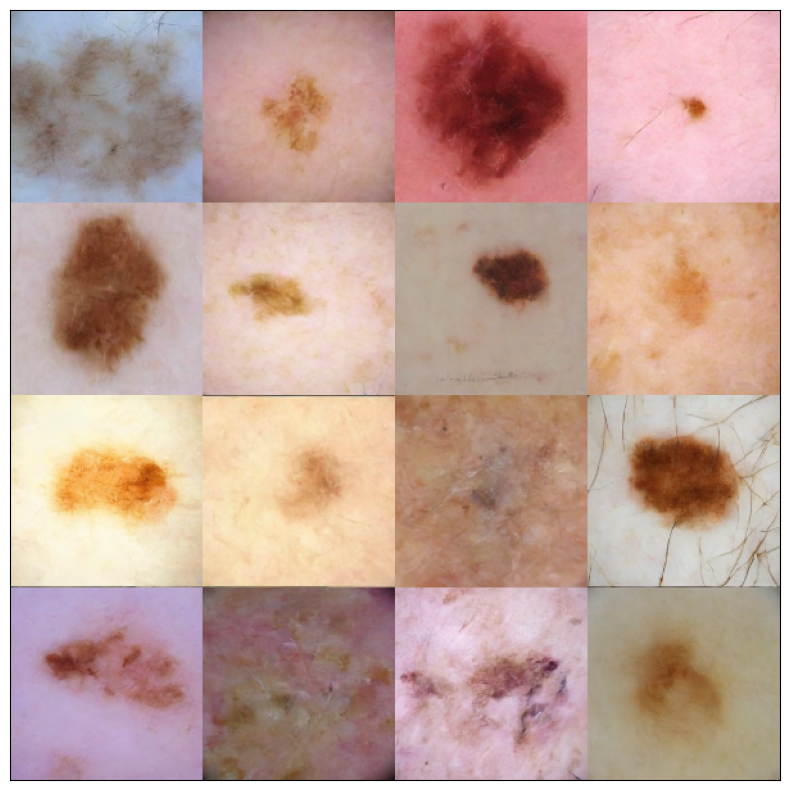}\\
        \Large{\textbf{I (a)}} & \Large{\textbf{I (b)}} & \Large{\textbf{I (c)}} & \Large{\textbf{I (d)}} & \Large{\textbf{I (e)}} & \Large{\textbf{I (f)}} \\

    \end{tabular}
}
    \resizebox{\linewidth}{!}{

    \begin{tabular}{ccccc} 
            \includegraphics[width=0.4\linewidth]{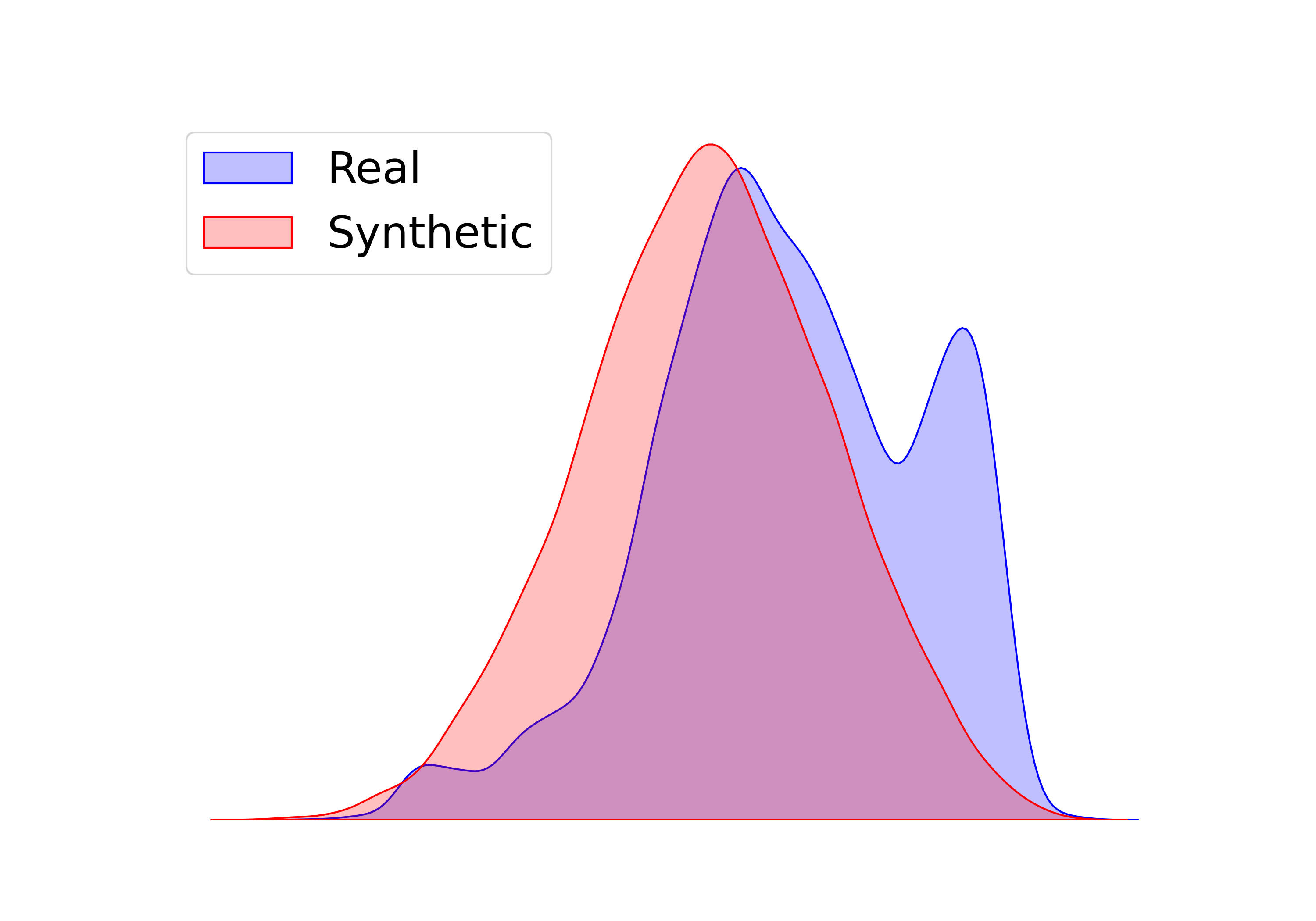} &
            \includegraphics[width=0.4\linewidth]{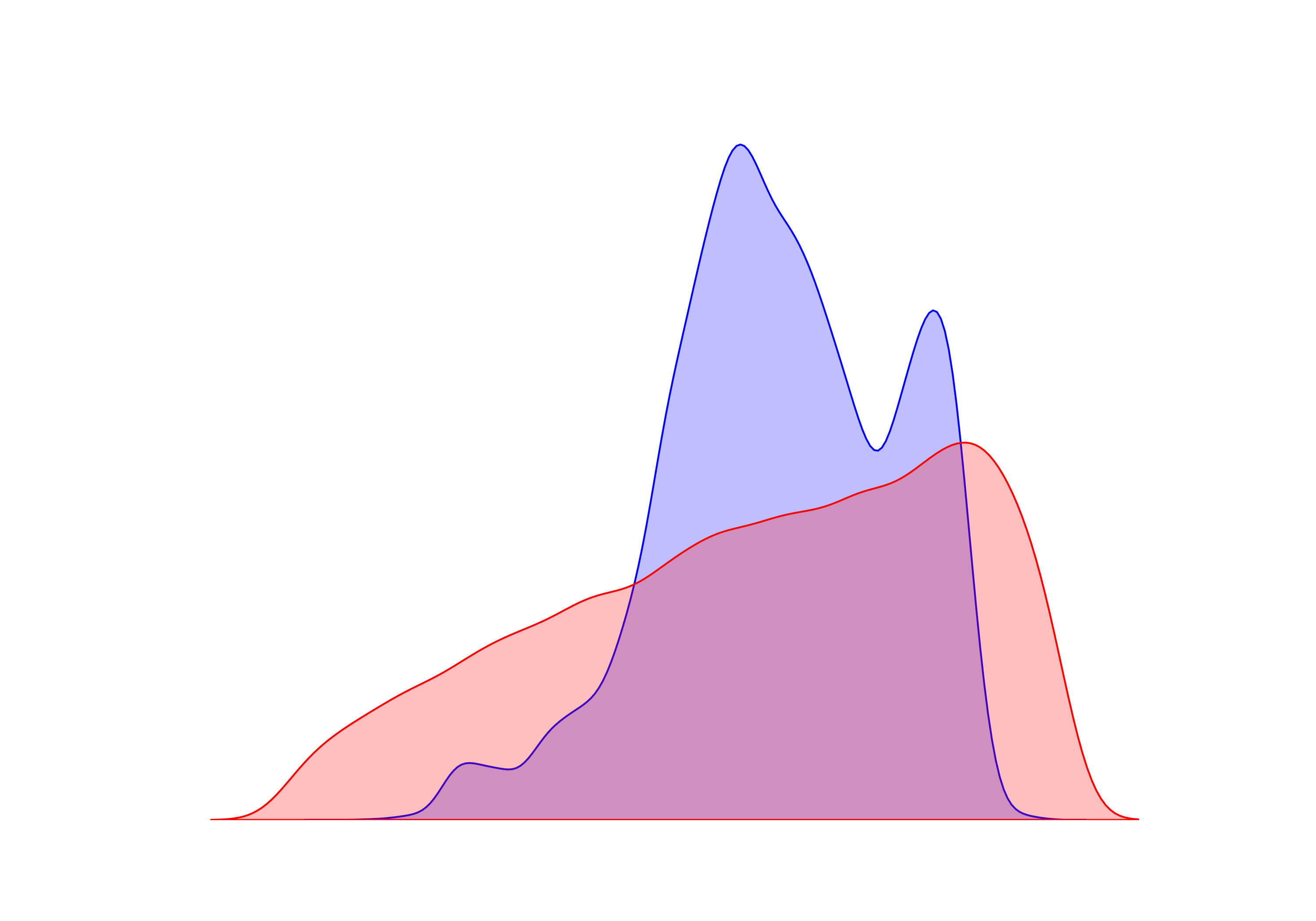} &
            \includegraphics[width=0.4\linewidth]{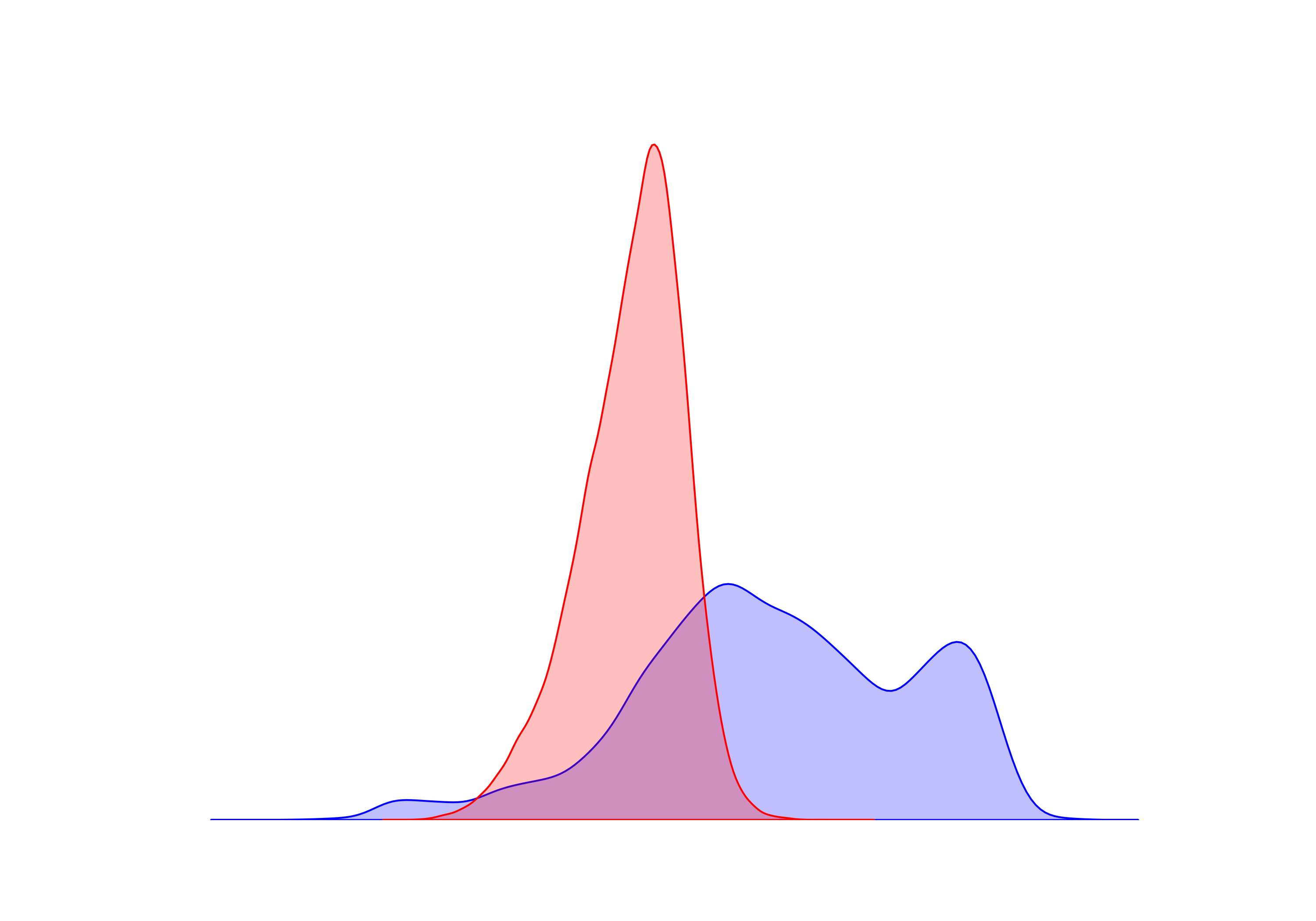} &
            \includegraphics[width=0.4\linewidth]{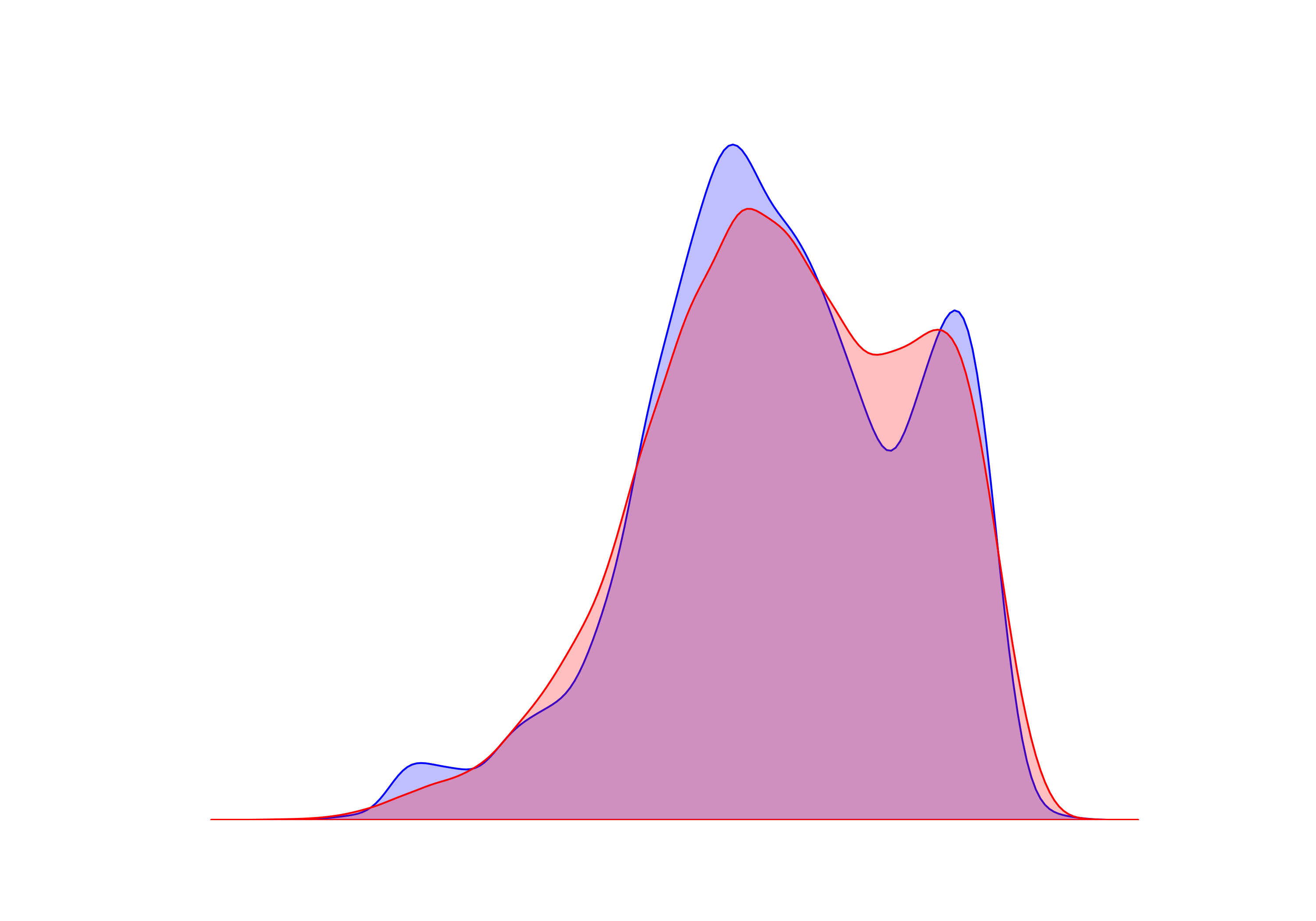}  & 
            \includegraphics[width=0.4\linewidth]{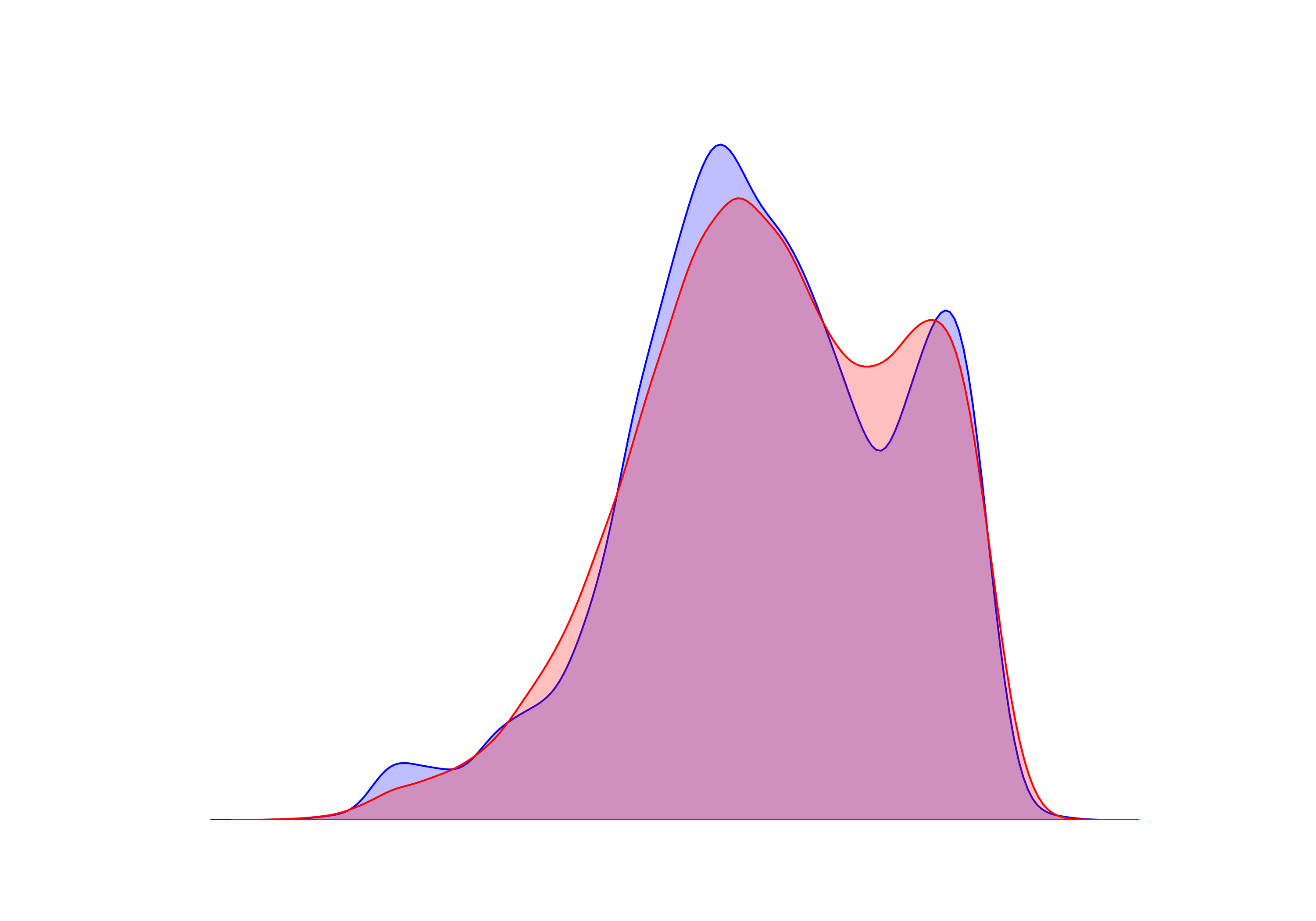} \\
            \Large{\textbf{II (a)}} & \Large{\textbf{II (b)}} & \Large{\textbf{II (c)}} & \Large{\textbf{II (d)}} & \Large{\textbf{II (e)}}\\
        \end{tabular}
    }

    \resizebox{\linewidth}{!}{

    \begin{tabular}{ccccc} 
        \includegraphics[width=0.4\linewidth]{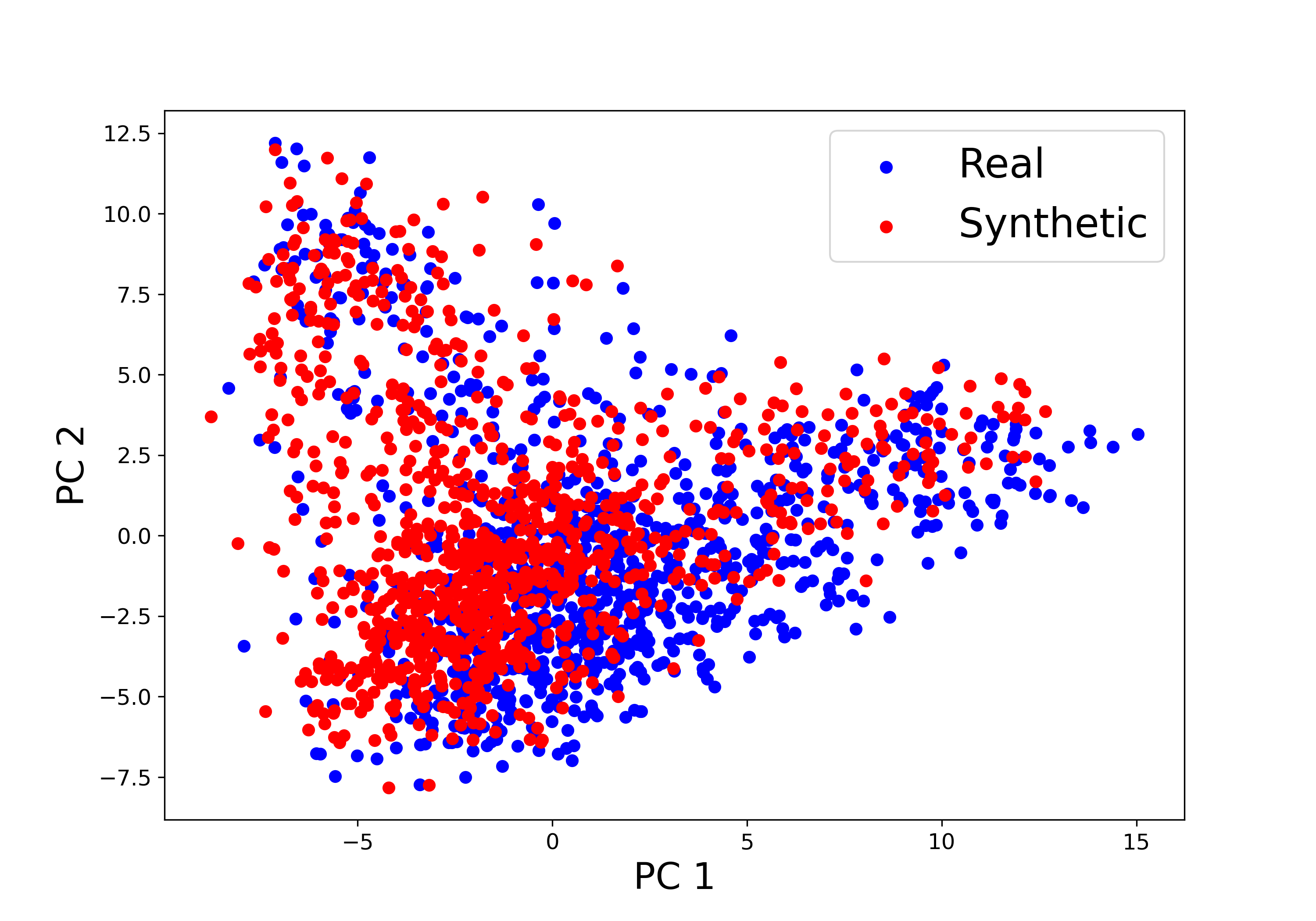} &
        \includegraphics[width=0.4\linewidth]{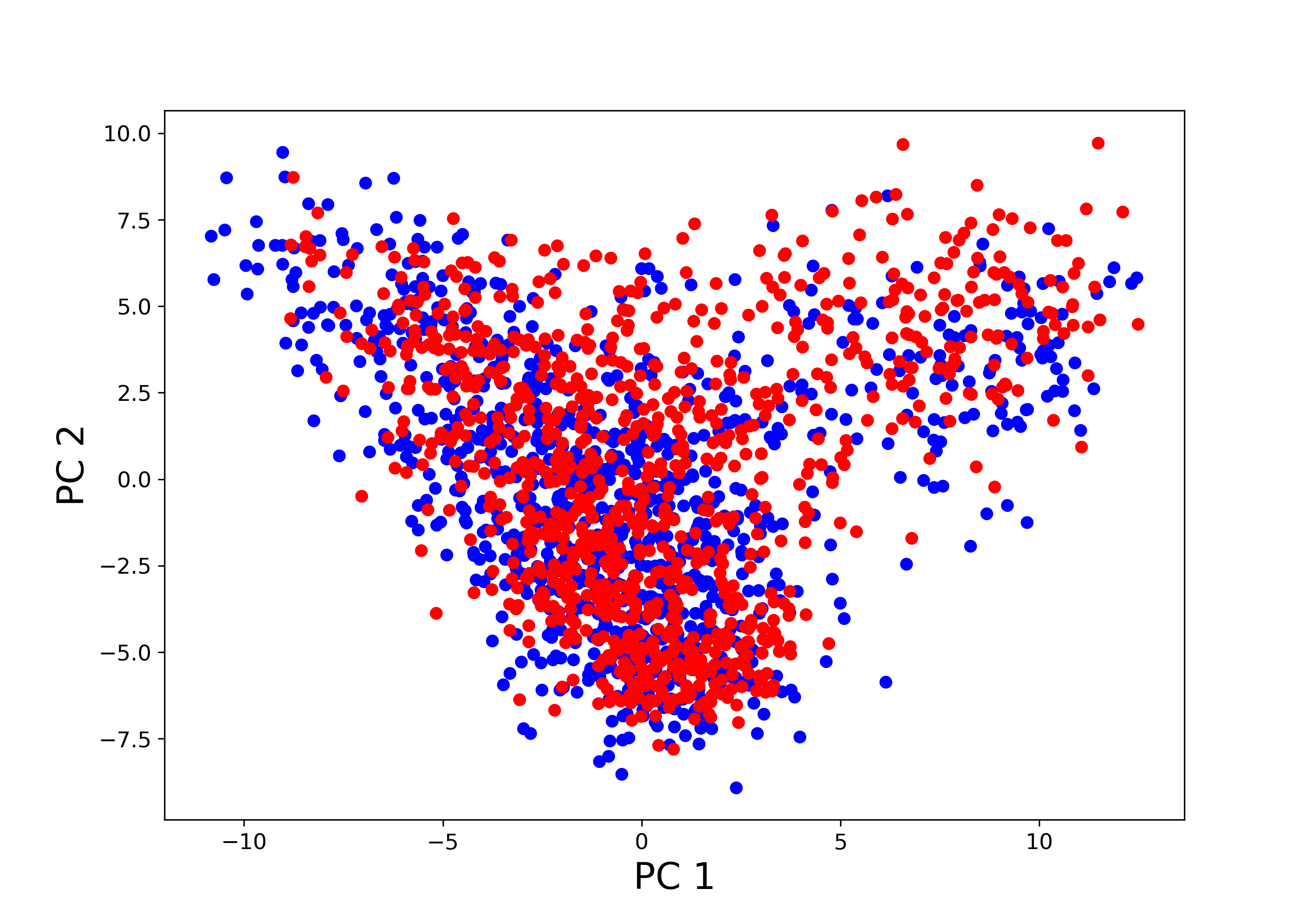} &
        \includegraphics[width=0.4\linewidth]{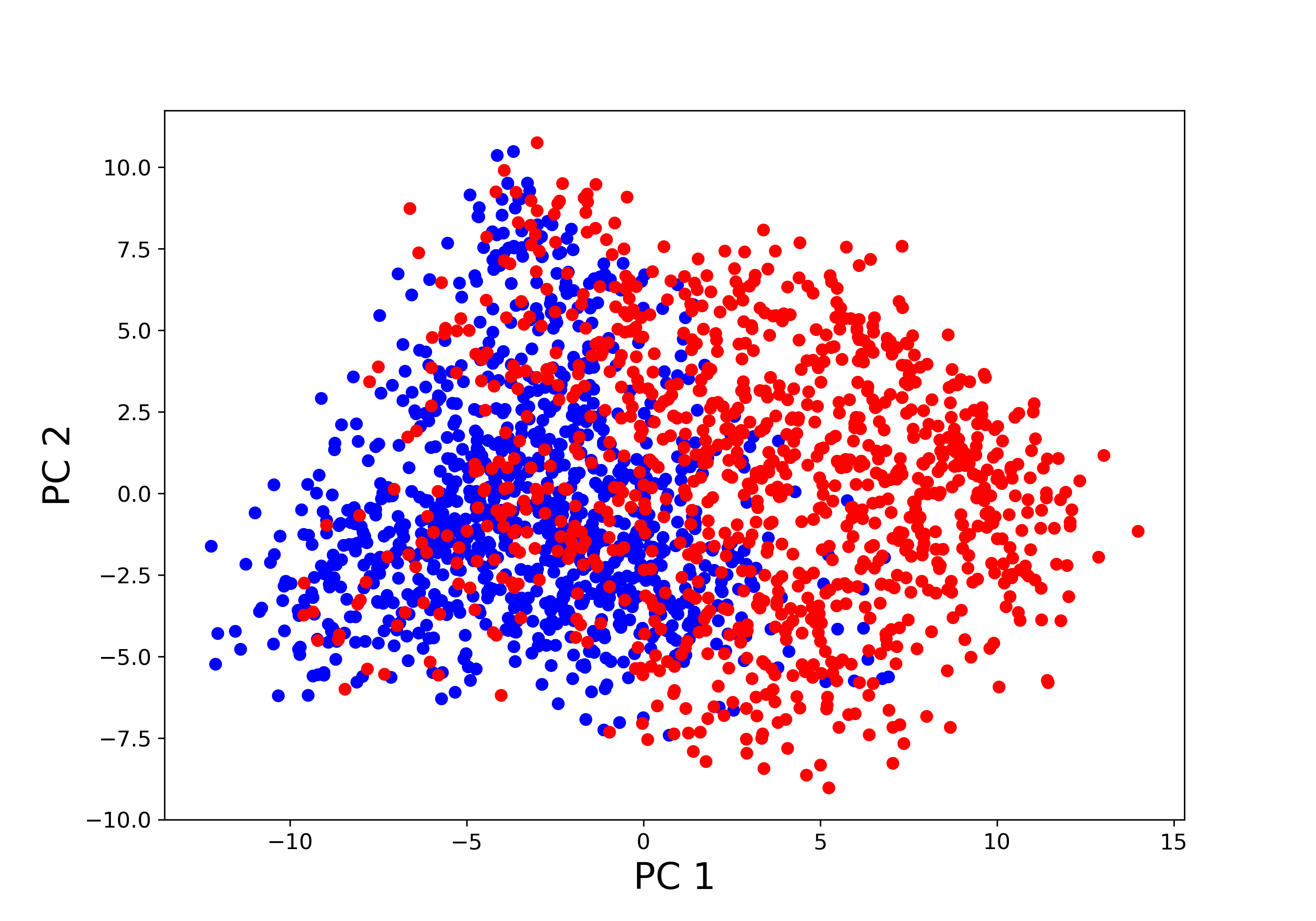} &
        \includegraphics[width=0.4\linewidth]{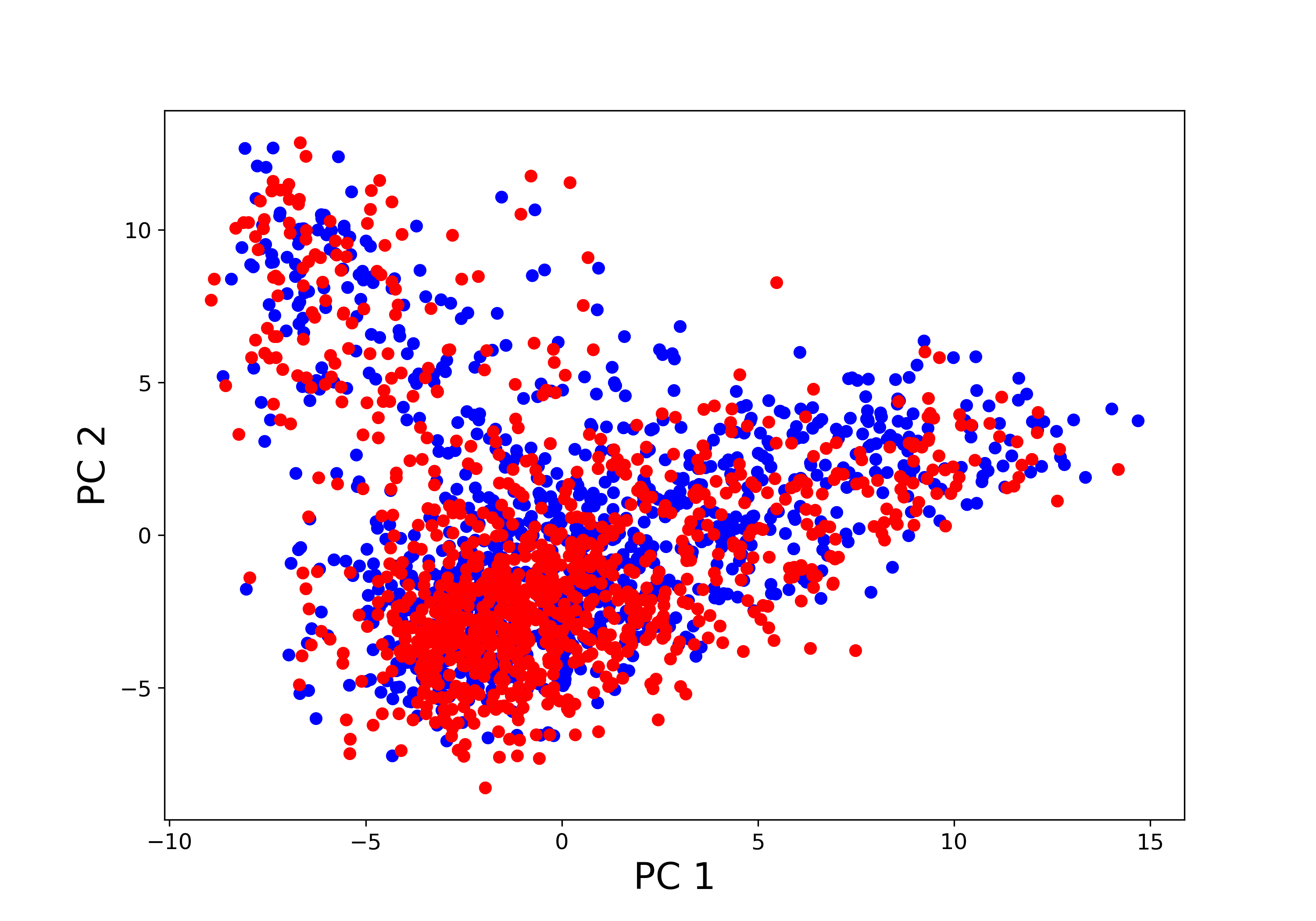}&
        \includegraphics[width=0.4\linewidth]{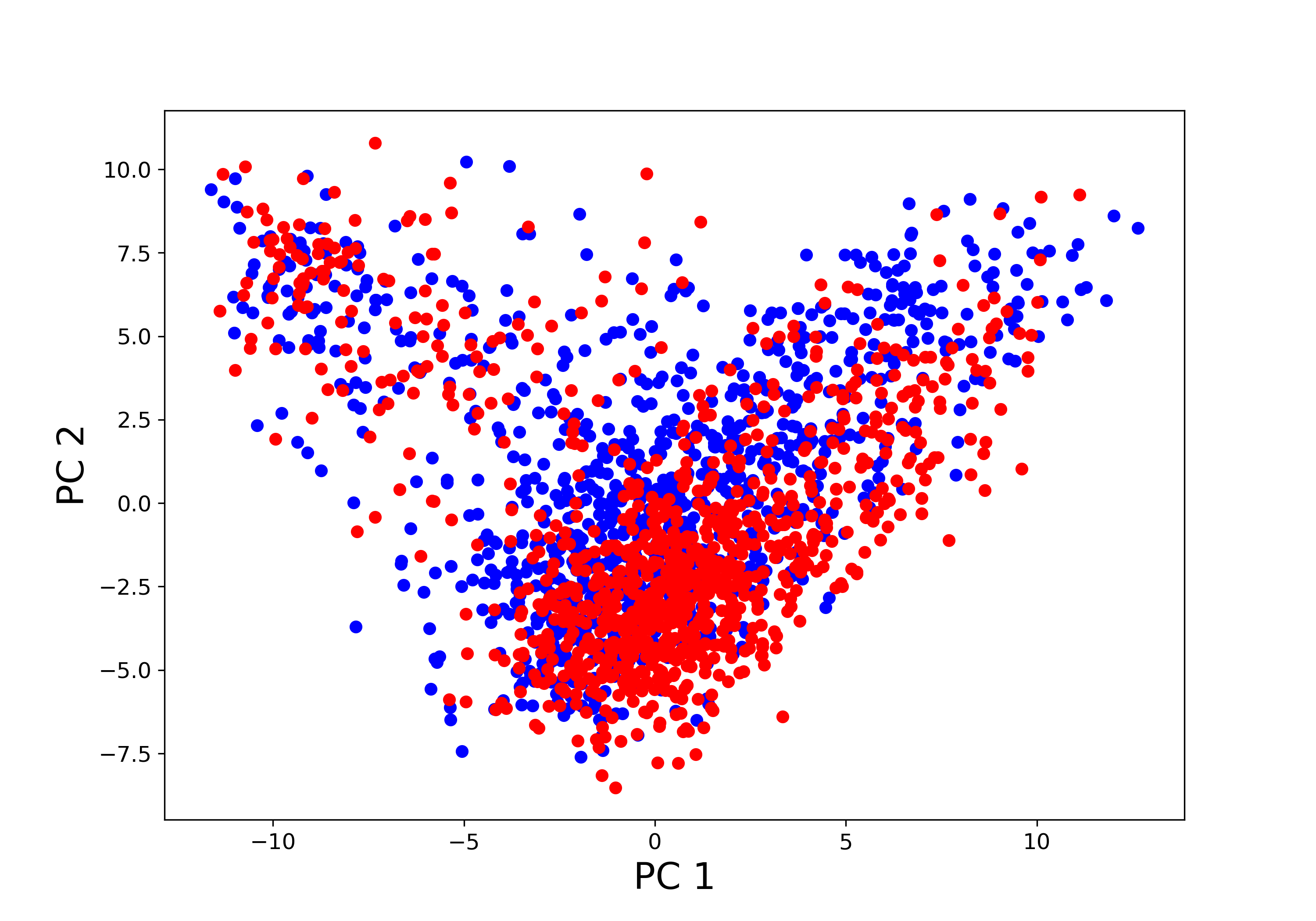}\\
        \Large{\textbf{III (a)}}
& \Large{\textbf{III (b)}} & \Large{\textbf{III (c)}} & \Large{\textbf{III (d)}} & \Large{\textbf{III (e)}}
    \end{tabular}
}
       
\caption{Sample image visualization from I.(a) ISIC real dataset and I. (b-f) synthetic images generated by the diffusion models. Visualization of the density distribution plot (II) and Principal Component Analysis (III) to compare the real ISIC data and the synthetic data generated by (a) Unconditional LDM, (b) Unconditional DiT, (c) LDM conditioned on LLaVA generated text prompt, (d) DiT conditioned on LLaVA generated text prompt, (e) DiT conditioned on LLaVA-Med generated text prompt.}
    \label{fig:all_graphs}
\end{figure*}

\noindent\textbf{Reverse process (Denoising):} The diffusion model learns the reverse process $p_\theta(x_{t-1} | x_t)$, where a neural network predicts both the mean $\mu_\theta(x_t)$ and the variance $\Sigma_\theta(x_t)$ of the reverse process. Training of this reverse process is simplified by reparameterizing the mean $\mu_\theta$ as a noise prediction by network $\epsilon_\theta$. The model is trained using the mean-squared error (MSE) between the predicted noise $\epsilon_\theta(x_t)$ and the true noise $\epsilon_t$ (sampled from a standard Gaussian distribution):
\begin{equation}
   \mathcal{L}_{\text{simple}}(\theta) = \|\epsilon_\theta(x_t) - \epsilon_t\|_2^2. 
\end{equation}
Once the model is trained, it can generate new data by initializing from random noise $x_{\text{max}} \sim \mathcal{N}(0, I)$ and iteratively transforming this noise using the learned reverse process. The diffusion process is performed on a compressed latent space rather than a high-dimensional pixel space. This process first encodes input dermoscopic images into latent representations $ z = E(x) $ using an encoder $( E )$, then trains a diffusion model on this latent space. A learned decoder is used to reconstruct the original dermoscopic images: $ x = D(z) $. We use the pre-trained variational autoencoder (VAE) model \cite{vae_enc} from the Stable Diffusion model \cite{stable_diff}. The Latent Diffusion Model (LDM) \cite{stable_diff} uses a U-Net architecture, and DiT \cite{dit} uses transformer-based architecture for the reverse process. We follow the DiT \cite{dit} model architecture and include DiT Blocks with a multi-head cross-attention mechanism. Similar to DiT \cite{dit}, our model applies patchify to the noisy image $z$ in the latent space and passes through several DiT blocks. Then the noise timestep $t$ and the text embedding $E_t$ are concatenated and passed to the multihead cross-attention layer of the DiT block for conditioning. The DiT block takes the noise input $z$ and applies multi-head self-attention on $z$, as shown in the DiT Block of Fig.~\ref{fig:arch1}. In our DermDiT model, the DiT block repeats 24 times, and we use a patch size of 4.
Finally, the DermDiT model predicts the sample noise $\tilde{\epsilon}$ from the text embedding $E_t$ and image representations with added noise $z_t$. 
\begin{equation}
    \epsilon_{\theta}(x_t) = {DiT}(VAE(x_t)\oplus\epsilon, t, E_t).
\end{equation}
\subsection{Image Generation}
To generate a new dermoscopic image, first, we select the attributes that we want in our image, such as benign disease, dark skin tone, etc. We create a text prompt including these attributes or we can use one of the text prompts generated by VLM that includes these attributes. The text prompt is then converted into a text embedding via the CLIP text encoder. A random noise $x_{t_{max}} \sim \mathcal{N}(0, I)$ is sampled from the normal distribution and passed to the DiT block with the text embedding to sample $ x_{t-1} \sim p_\theta (x_{t-1} \mid x_t) $. We use $t_{max}$ = 250 sampling steps, similar to DiT \cite{dit}. After sampling $x_0$, we use a VAE decoder to convert it into a dermoscopic image.

\section{Experimental Evaluation}
\label{sec:evaluation}

\subsection{Implementation Details}

\textbf{Data:} The International Skin Imaging Collaboration (ISIC) datasets [2016-2020] \cite{isic2018skin, tschandl2018ham10000, isic2020} are the most popular dataset for skin disease diagnosis. They comprise thousands of dermoscopic images in addition to gold-standard disease diagnostic metadata. Although their metadata do not include any skin tone information, they contain other attributes such as sex, approximate age, etc.  The Diverse Dermatology Images (DDI) dataset \cite{ddi} is a diverse dataset compared to any other dataset. It contains a total of 656 dermatology images of three different skin tone categories based on Fitzpatrick skin types (FST) \cite{fitzpatrick1988validity}. We define FST I-II (light skin tone), FST III-IV, and FST V-VI(dark skin tone) as the skin tone type A, B, and C, respectively, to standardize classification and support analysis across diverse skin tones. The Fitzpatrick17k dataset \cite{groh2021fitzpatrick} contains clinical images of skin conditions, including skin tone information, Fitzpatrick skin types (FST).

\noindent\textbf{Inputs:} The input dermoscopic images are resized to $256\times256$ resolution and 0--1 normalized.  

\noindent\textbf{Training:} We follow DermDiff \cite{munia2025dermdiff}  to get skin tone information for ISIC. We train our DermDiT generative model with the ISIC dataset. Following DermDiff \cite{munia2025dermdiff}, we also build a ResNeXt-based skin diagnosis model for downstream classification evaluation. We implemented our models in Python with the Pytorch library and executed them on an Intel(R) Xeon(R) 128GB machine with two NVIDIA RTX A4000 GPUs.

\noindent\textbf{Hyper-parameters:} We trained our generative model with a mini-batch size of 16 and a learning rate of $1e^{-4}$ for 400k steps. For classification, we trained with a mini-batch size of 64 for 10 epochs with ISIC datasets and took the model with the lowest validation loss. 

\noindent\textbf{Evaluation:} For image generation, we calculate Fr\'echet Inception Distance (FID) \cite{fid}and multi-scale structural similarity index metric (MS-SSIM) scores. For classification models, we report the F1-score and AUC scores.

\begin{table}
\centering
\caption{Performance comparison of the proposed DermDiT against the baseline LDM and DiT models trained on the ISIC dataset. FID and MS-SSIM scores are reported for unconditional and conditional (varying attributes) image generation.}
\resizebox{\linewidth}{!}{
\Large
\begin{tabular}{@{}lllcccc@{}}
\toprule
\multirow{2}{*}{Data} & \multirow{2}{*}{VLM} & \multirow{2}{*}{Model} & FID & FID & MS-SSIM \\ 
&&& (Pre-trained) & (Fine-tuned) && \\
\midrule
ISIC & - & - & - & - & 0.445 \\
\midrule
\multirow{2}{*}{ISIC (No condition)} & \multirow{2}{*}{-} & LDM & 39.25 & 52.94 & 0.458 \\
 && DiT & 67.53 & 24.36 & 0.418 \\
 \midrule
\multirow{2}{*}{ISIC (2 attributes)} & \multirow{2}{*}{LLaVA} & LDM & 135.68 & 269.86 & 0.511 \\
 && DiT & \textbf{30.37} & 6.08 & 0.522 \\
  \midrule
ISIC (5 attributes) & LLaVA-Med & DiT & 38.62 & \textbf{5.75} & \textbf{0.438} \\
\bottomrule
\end{tabular}
}
\label{table:fid}
\end{table}

\subsection{Results and Discussion}
\paragraph*{Image Generation Performance:}

First, we trained the LDM and DiT models on the ISIC dataset without any metadata, treating them as unconditional diffusion models. Their generative performance is summarized in Table \ref{table:fid}. Next, we selected two attributes from the ISIC metadata: (a) disease type (benign or malignant) and (b) skin tone (light, brown, dark), with skin tone information sourced from \cite{munia2025dermdiff}. Following Section 2.1, we used the LLaVA model to generate text prompts incorporating these attributes, subsequently training the DiT and LDM models with these prompts. Additionally, we extended to five attributes: (a) disease type, (b) skin tone, (c) gender, (d) age, and (e) anatomical site, creating captions for these attributes using the LLaVA-Med model. We generated 30k samples for unconditional models and 50k samples for conditional models.

\noindent We evaluated our DermDiT model's fidelity using the FID score, which measures the similarity between the distributions of real and synthetic data. Lower FID scores indicate that the generated images closely resemble the distribution of the real images. Although the FID score traditionally uses an Inception-v3 model \cite{fid} trained on ImageNet, we fine-tuned the Inception-v3 model for disease classification on our dermatology dataset to provide a more accurate representation for our evaluation. Table \ref{table:fid} shows the FID scores compared with real training data with both pre-trained and fine-tuned Inception-v3 models. We also calculated MS-SSIM to evaluate the diversity of these models. But first, we calculated the MS-SSIM score for the real ISIC dataset to compare the diversity of the training dataset and the generative data. From Table \ref{table:fid}, we can see that the DiT model trained with LLaVA-Med generated captions with five attributes are more diverse than the real training data. Also this model has the lowest FID score when calculated with the fine-tuned Inception-v3 model. 
We also visualize the data distribution of the generated samples via Density plots in Fig. \ref{fig:all_graphs}. The density distribution of the synthetic images generated by the DiT model trained with LLaVA and LLaVA-Med generated prompts quite overlapping the real ISIC data distribution (Fig. \ref{fig:all_graphs} II(d) and II(e)). We also visualize the Principal Component Analysis (PCA) by randomly sampling 1000 images from the dataset and getting their image embedding from the Inception-v3 model. From the image embeddings, we apply PCA to get two Principal Components (PC1 and PC2) and plot their distribution. Fig. \ref{fig:all_graphs} demonstrates that synthetic images generated by the DiT model overlap with the training data, but for LDM, their distribution is more separable. We show some sample images from all the models in the first row of Fig. \ref{fig:all_graphs}.

\begin{table}[t]
\centering
\caption{Quantitative evaluation of real and synthetic data models by measuring AUC, Recall, and F1 scores when tested on 9 different test datasets.}
\label{tab:skin_disease_results}
\resizebox{0.9\linewidth}{!}{%
\begin{tabular}{@{}ll c c c c c@{}}
\toprule
Test data & Training data & \phantom{a} & AUC & Recall & F1-score \\ 

\midrule
\multirow{2}{*}{DDI} & Real && 0.587 & 0.070 & 0.128  \\
& Synthetic && 0.527 & 0.632 & 0.379  \\

\midrule
\multirow{2}{*}{DDI (A)} & Real &&  0.546 & 0.041 & 0.078   \\
& Synthetic && 0.419 & 0.653 & 0.322  \\

\midrule
\multirow{2}{*}{DDI (B)} & Real &&  0.651 & 0.081 & 0.148   \\
& Synthetic &&  0.593 & 0.649 & 0.453 \\

\midrule
\multirow{2}{*}{DDI (C)} & Real && 0.538 & 0.083 & 0.143    \\
& Synthetic &&  0.544 & 0.583 & 0.352  \\

\midrule
\multirow{2}{*}{ISIC-2018} & Real && 0.870 & 0.211 & 0.327    \\
& Synthetic &&  0.715 & 0.848 & 0.299   \\

\midrule
\multirow{2}{*}{Fitzpatrick17k} & Real && 0.605 & 0.078 & 0.141    \\
& Synthetic &&  0.582 & 0.421 & 0.496    \\

\midrule
\multirow{2}{*}{Atlas Derm} & Real && 0.811 & 0.238 & 0.359  \\
& Synthetic &&  0.612 & 0.675 & 0.420  \\

\midrule
\multirow{2}{*}{Atlas Clinical} & Real && 0.667 & 0.107 & 0.187  \\
& Synthetic &&  0.562 & 0.599 & 0.383  \\

\midrule
\multirow{2}{*}{ASAN} & Real &&  0.853 & 0.153 & 0.202 \\
& Synthetic && 0.594 & 0.322 & 0.136  \\

\midrule
\multirow{2}{*}{MClass Derm} & Real && 0.842 & 0.250 & 0.370  \\
& Synthetic &&  0.704 & 0.850 & 0.430 \\

\midrule
\multirow{2}{*}{MClass Clinical} & Real && 0.822 & 0.250 & 0.333\\
& Synthetic &&  0.756 & 0.350 & 0.412 \\

\bottomrule
\end{tabular}%
}
\end{table}

\paragraph*{Downstream Diagnostic Performance:}
We build a binary classification model with ResNeXt-101 architecture to predict the skin disease class: benign or malignant. We train this model with the ISIC real dataset and test the model on various test datasets. We evaluate both in-domain (ISIC-2018) and out-domain, such as clinical datasets (DDI and Fitzpatrick). We also considered testing them on other publicly available datasets such as Atlas \cite{atlas2018seven}, ASAN \cite{asan_dataset}, and MClass \cite{mclass_data} datasets.  We also train this classification model with only synthetic data generated by the DiT model trained with LLaVA-Med text prompts. As we have seen this model is able to generate better and more realistic images than other models. We generate synthetic images using this model to create a balanced dataset where all the attributes have similar data distribution. Then we train the same ResNeXt model with this newly generated synthetic data. We evaluate the model on the same test datasets and show their AUC, Recall, and F1-score in Table \ref{tab:skin_disease_results}. Although the synthetic images didn't improve in terms of AUC, it has better Recall and F1 scores than the model trained with real ISIC images. Even though the model is trained with only synthetic images, its AUC of over fifty percent and improved F1-score demonstrate that the quality of the synthetic images is sufficient to capture meaningful patterns and support reliable model performance of the DermDiT model.

\section{Conclusions}
Our proposed DermDiT model focuses on generating synthetic images for under-represented subgroups based on given attributes only. By leveraging vision-language models, we can incorporate descriptive information into text prompts, enhancing the robustness and generalizability of the generative model. The synthetic images not only closely resemble real data but also demonstrate higher diversity than the real ISIC dataset. This approach can potentially reduce reliance on real data, supporting applications where real medical data is limited while still ensuring effective and reliable diagnostic performance.

\section{Acknowledgments}
\label{sec:acknowledgments}
This work was funded by the UNITE Research Priority Area at the University of Kentucky.

\balance
\bibliographystyle{IEEEbib}
\bibliography{refs}

\end{document}